\documentclass[11pt]{article}

\usepackage{multirow}
\usepackage[table]{xcolor}
\usepackage{coling2020}
\colingfinalcopy

\usepackage{times}
\usepackage{url}
\usepackage{latexsym}

\usepackage{amssymb}
\usepackage{amsmath}
\usepackage{bbm}
\usepackage{array}
\usepackage{graphicx}
\usepackage{arydshln}
\usepackage{booktabs}

\usepackage[labelfont=bf]{caption}
\usepackage{subcaption}

\usepackage{bussproofs}

\usepackage{tikz}
\usetikzlibrary{patterns,shapes,arrows,positioning,calc}
\usepackage{pgfplots} 

\usepackage{amsmath,amsfonts,bm}

\def\eqref#1{equation~\ref{#1}}

\def\1{\bm{1}}

\def\mS{{\bm{S}}}

\DeclareMathAlphabet{\mathsfit}{\encodingdefault}{\sfdefault}{m}{sl}
\SetMathAlphabet{\mathsfit}{bold}{\encodingdefault}{\sfdefault}{bx}{n}
\newcommand{\tens}[1]{\bm{\mathsfit{#1}}}

\def\tS{{\tens{S}}}

\def\tW{{\tens{W}}}

\def\emS{{S}}

\newcommand{\etens}[1]{\mathsfit{#1}}

\def\etS{{\etens{S}}}

\def\etW{{\etens{W}}}

\newcommand{\ruleref}[1]{{\bf(\subref{#1})}}

\newcommand{\bert}{\textsc{Bert}}
\newcommand{\tiger}{\textsc{Tiger}}
\newcommand{\negra}{\textsc{Negra}}
\newcommand{\dptb}{\textsc{DiscPTB}}

\title{%
	Span-based discontinuous constituency parsing:\\
	a family of exact chart-based algorithms \\
	with time complexities from $\mathcal O(n^6)$ down to $\mathcal O(n^3)$ 
}

\author{Caio Corro\thanks{\hspace{0.2cm}Work partially done while the author was a postdoc at University of Amsterdam with Ivan Titov.} \\
  Université Paris-Saclay, CNRS, LIMSI, 91400, Orsay, France.%
  \\
  {\tt caio.corro@limsi.fr} \\
}

\date{}

\begin{document}
\maketitle
\begin{abstract}
We introduce a novel chart-based algorithm for span-based parsing of discontinuous constituency trees of block degree two, including ill-nested structures.
In particular, we show that we can build variants of our parser with smaller search spaces and time complexities ranging from $\mathcal O(n^6)$ down to $\mathcal O(n^3)$.
The cubic time variant covers 98\% of constituents observed in linguistic treebanks while having the same complexity as continuous constituency parsers.
We evaluate our approach on German and English treebanks (\negra{}, \tiger{} and \dptb{}) and report state-of-the-art results in the fully supervised setting.
We also experiment with pre-trained word embeddings and \bert{}-based neural networks.
\end{abstract}

\section{Introduction}

Syntactic parsing aims to recover the latent syntactic relations between words in a sentence, expressed in a given syntactic formalism.
In this paper, we focus on constituency trees where the syntactic structure is described by the means of a hierarchical structure composed of nodes:
words are leaf nodes whereas internal nodes represent labeled constituents or phrases, see Figure~\ref{fig:example}.
Constituency trees can broadly be classified into two categories.
On the one hand, in a \emph{continuous} constituent tree, each node must dominate a contiguous sequence of words.\footnote{The set of words that a node dominates is the set of leaf nodes in the subtree for which this node is the root.}
On the other hand, in a \emph{discontinuous} constituent tree, a node can dominate a non-contiguous sequence of words.
It has been argued that modeling discontinuity is unavoidable, see for example \newcite{mccawley1982discontinuous} and \newcite{bunt1987discontinuous} for English and \newcite{muller2004continuous} for German.

Phrase-structure grammars have been proposed to parse and generate constituency trees.
For example, Context-Free Grammars (CFG) are able to process continuous constituent trees whereas Multiple Context Free Grammars \cite[MCFG]{seki1991mcfg} and Linear Context-Free Rewriting System \cite[LCFRS]{vijay-shanker1987lcfrs} are able to process their discontinuous counterpart.
CFGs have been widely studied for practical parsing due to the availability of time-efficient parsing algorithms based on \emph{chart-based algorithms} (i.e.\ dynamic programming): parsing a sentence of length $n$ is a $\mathcal O(gn^3)$ problem where $g$ is a grammar related constant \cite{kasami1966cyk,younger1967cyk,Cocke1969cyk}.
However, parsing algorithms for MCFGs and LCFRSs are deemed to be impractical despite their polynomial-time complexity (see Section~\ref{background:lcfrs}).
Therefore, most of the experimental work in this field has been limited to parsing short sentences, e.g.\ sentences that contains less than 40 words \cite{kallmeyer2010lcfrs,evang2011plcfrs,maier2012plcfrs,kuhlmann2006mildly}.

Advances in machine learning led to the development of constituency parsers that are not based on phrase-structure grammars.
Instead, the prediction step only ensures the well-formedness of the resulting structure and does not enforce compliance of the syntactic content represented by the structure.
For example, a verbal phrase is not constrained to contain a verb.
As such, they can be assimilated to the mainstream approach to bi-lexical dependency parsing
where one consider candidate outputs only in a restricted class of graphs: non-projective \cite{mcdonald2005mst}, projective \cite{eisner1997bilexical} or bounded block degree and well-nested spanning aborescences \cite{gomez2009nonproj,gomez2011nonproj,corro2016bdwn}, among others \cite{kuhlmann2006mildly,satta2013headsplit,pitler2012gapminding}.
These approaches assume that intricate relations in the syntactic content can be implicitly learned by the scoring function.

Span-based parsing is a grammarless approach to constituency parsing that decomposes the score of a tree solely into the score of its constituents, originally proposed for continuous constituency parsing \cite{hall2014spanbased,stern2017spanbased,cross2016span}.%
\footnote{%
In contrast, for example, to several transition systems that can incorporate scores related to actions that where executed during derivation,
or to split point decision and left-right span scores in the parser of \newcite{stern2017spanbased}.
}
Recovering the tree of highest score can be done exactly using a slightly updated CYK algorithm or using inexact\footnote{The term inexact refers to the fact that these methods are no guaranteed to recover the highest scoring structure.} methods like top-down or transition based algorithms.
This approach has obtained state-of-the art results for continuous constituency parsing \cite{stern2017spanbased,kitaev2018parser,kitaev2019bertparser}.
In this work, we propose the first span-based parser with an exact decoding algorithm for discontinuous constituent parsing.
To this end, we introduce a novel exact chart-based algorithm based on the parsing-as-deduction formalism \cite{pereira1983deduction} that can parse constituent trees with a block degree of two,
including ill-nested structures (see Section~\ref{sec:alg}).
which have been argued to be unavoidable to model natural languages \cite{chen2010illnestedness}.
Despite its $\mathcal O(n^6)$ time-complexity, where $n$ is length of the input sentence, the algorithm is reasonably fast: all treebanks can be parsed without removing long sentences.
Moreover, we observe that several deduction rules are of little use to retrieve trees present in treebanks.
Therefore, we experiment with variants of the algorithm where we remove specific deduction rules.
This leads to parsing algorithms with lower asymptotic complexity that experimentally produce accurate parses.
Importantly, we show that a specific form of discontinuity can be parsed in $\mathcal O(n^3)$, that is with the same asymptotic complexity as continuous constituency parsing.

Even with a constraint on the block degree, there are $\mathcal O(n^4)$ prospective constituents that all have to be scored as we rely on exact decoding without further assumption and/or filtering.
This would be too expensive in practice.%
\footnote{For example, in preliminary experiments we found that the neural network computing scores could not fit on a 12GB GPU.}
Transition-based models address this problem by only scoring constituents that are requested during beam search.
Although this is appealing on CPU, this lazy computation of scores cannot fully benefit from modern GPU architectures to parallelize computation at test-time.
In this work, we propose to decompose the score of a constituent into independent parts leading to a quadratic number of scores to compute.
As such, we can rely on efficient dimension broadcasting and operation batching available on modern GPUs.

\begin{figure*}
	\begin{minipage}{.55\textwidth}
			\begin{tikzpicture}[word/.style={font=\small}, cst/.style={font=\scriptsize, draw}]
		\node (wstart) {};
		\node[word,red ,right=of wstart, xshift=-0.5cm] (w0) {\bf What};
		\node[word,right=of w0,xshift=-0.5cm] (w1) {I};
		\node[word,right=of w1,xshift=-0.5cm] (w2) {said};
		\node[word,right=of w2,xshift=-0.5cm] (w3) {should};
		\node[word,right=of w3,xshift=-0.5cm] (w4) {I};
		\node[word,right=of w4,xshift=-0.5cm, red] (w5) {\bf do};
		\node[right=of w5,xshift=-0.5cm, red] (wend) {};
		
		\node[right=of w0, xshift=-1.2cm] {,};
		\node[right=of w2, xshift=-1.2cm] {,};
		\node[right=of w5, xshift=-1.2cm] {?};
		
		\node[font=\scriptsize] at ([yshift=-0.25cm]$(wstart)!0.5!(w0)$) {0};
		\node[font=\scriptsize] at ([yshift=-0.25cm]$(w0)!0.5!(w1)$) {1};
		\node[font=\scriptsize] at ([yshift=-0.25cm]$(w1)!0.5!(w2)$) {2};
		\node[font=\scriptsize] at ([yshift=-0.25cm]$(w2)!0.5!(w3)$) {3};
		\node[font=\scriptsize] at ([yshift=-0.25cm]$(w3)!0.5!(w4)$) {4};
		\node[font=\scriptsize] at ([yshift=-0.25cm]$(w4)!0.5!(w5)$) {5};
		\node[font=\scriptsize] at ([yshift=-0.25cm]$(w5)!0.5!(wend)$) {6};
		
		\node[cst] (np_I1) at ([yshift=0.75cm]w1) {NP};
		
		\node[cst] (vp) at ([yshift=2.75cm]w2) {VP};
		\node[cst] (s) at ([yshift=3.5cm]w2)  {S};
		
		\node[cst] (np_I2) at ([yshift=0.75cm]w4) {NP};
		\node[cst] (sq) at ([yshift=2cm]w4) {SQ};
		\node[cst] (sbarq) at ([yshift=2.75cm]w4) {SBARQ};

		\node[cst, line width=0.5mm,red] (vp_do) at ([yshift=1.25cm]w5) {VP};
		\node[cst, line width=0.5mm,red] (whnp) at ([yshift=0.75cm]w0) {WHNP};
		\draw[line width=0.5mm,red] (vp_do) -| (whnp);
		\draw[line width=0.5mm,red] (vp_do) -- (w5);
		\draw[line width=0.5mm,red] (whnp) -- (w0);

		\draw (np_I1) -- (w1);
		\draw (vp) -- (w2);
		\draw (np_I2) -- (w4);
		
		\draw (sbarq) -- (sq);
		\draw (sq) -| (w3);
		\draw (sq) -- (np_I2);
		\draw (sq) -| (vp_do);
		
		\draw (s) -- (vp);
		\draw (vp) -- (sbarq);
		\draw (vp) -| (np_I1);
	\end{tikzpicture}
		\captionof{figure}{%
			Exemple of discontinuous constituency tree.
			The bold red VP node dominates two sequences of words: "What" and "do".
			All other nodes are continuous.
			Numbers below the sentence are interstice indices used in the algorithm description.
		}
		\label{fig:example}
	\end{minipage}\hfill%
	\begin{minipage}{.4\textwidth}
		\begin{tikzpicture}
		\begin{axis}[
		width=6cm,
		height=5cm,
		xlabel={Sentence length},
		ylabel={Seconds per sentence},
		xmin=0, xmax=150,
		ymin=0, ymax=1.5,
		tick style={color=white},
		xtick={50,100,150},
		ytick={0, 0.5, 1.0,1.5},
		legend pos=north west,
		axis lines=left
		]
		
		\addplot[black, solid, thick] table[col sep=comma, x=x, y=y] {timing/src/pts_3};
		
		\addplot[black, dashed, thick] table[col sep=comma, x=x, y=y] {timing/src/pts_4};
		\legend{$\mathcal O(n^3)$, $\mathcal O(n^4)$};
		
		\end{axis}
		\end{tikzpicture}
		\captionof{figure}{%
			Execution time per sentence length of the chart-based algorithm for the $O(n^3)$ (solid line) and $O(n^4)$ (dashed lines) variants.
		}
		\label{fig:optimization}
	\end{minipage}
\end{figure*}

Our main contributions can be summarized as follows:
\begin{itemize}
    \item we propose a new span-based algorithm for parsing discontinuous constituency trees of block degree two with exact decoding and reasonable execution time;
    \item we propose a cubic-time algorithm that can parse a significant portion of discontinuous constituents in different corpora while having the same theoretical complexity as continuous constituency parsers;
    \item we report state-of-the-art parsing results on these treebanks in a fully supervised setting and experiment with pre-trained word embeddings, including \bert{} based models.
\end{itemize}
We release the C++/Python implementation of the parser.\footnote{\url{https://github.com/FilippoC/disc-span-parser-release}}

\section{Related Work}

{\bf Phrase-structure grammars:}
\label{background:lcfrs}
The LCRFS formalism has been widely used in the context of discontinuous constituency parsing, although MCFG and Simple Range Concatenation Grammars \cite{boullier1998proposal} have been shown to be equivalent, see \newcite{seki1991mcfg} and \newcite{boullier2004rcg}.
\newcite{kallmeyer2010lcfrs} introduced the first practical chart-based LCFRS parser for German, which was subsequently applied to English \cite{evang2011plcfrs}.
However, they restrict their data to sentences that contains fewer than 25 words.
To improve parsing time, \newcite{maier2012plcfrs} proposed an experimentally faster parser based on the $A^*$ search algorithm together with a block degree two restriction.
However, they still limit the sentence size to a maximum of 40 words.
A single sentence of 40 words takes around 3 minute to be parsed, an impressive improvement over the parser of \newcite{kallmeyer2010lcfrs} that needs several hours, but still prohibitively slow for large scale parsing.

{\bf Graph based parsing:}
A different line of work proposed to explore constituency parsing as a dependency parsing problem.%
\footnote{Note that opposite line of work also exists, that is reducing dependency parsing to constituency parsing, see for example \newcite{maier2010nonproj}.}
In other words, even if it is straightforward to represent constituency trees as hierarchical phrase structures,
the same syntactic content can be represented with different mathematical objects \cite{rambow2010opinion}, including directed graphs commonly used for dependency parsing.
\newcite{gonzalez2015reduction} reduced the (lexicalized) constituency parsing task to dependency parsing where the constituency structure is encoded into arc labels.
Then, discontinuous constituency parsing is reduced to the labeled \emph{Spanning Arborescence} problem which can be solved in quadratic time.
The same reduction has also been used in a sequence-to-sequence framework \cite{fernandez2020pointer}.
\newcite{corro2017gmsa} proposed a joint supertagging and dependency parsing reduction where vertices represents supertags\footnote{A supertag is an elementary tree that encodes the sequence of lexicalized constituents for which a given word is the head, see \newcite{bangalore1999supertagging}} and labeled arcs encode combination operations (substitution and adjunction).
The problem is then reduced to the labeled \emph{Generalized Spanning Arborescence} problem which is a known NP-hard optimization problem \cite{myung1995gmsa}.
One benefit of these approaches is that they do not assume any restriction on the constituency structure: they can parse ill-nested structures and have no block degree restriction.
However, they cannot impose such constraints which may be beneficial or required and they factor the score of a tree into dependency, supertag and/or label scores,
which means that the learning objective is not directly related to the evaluation metric which focuses on constituents.
Moreover, factorization rely on possibly erroneous heuristics (head-percolation tables) to lexicalize the original structure if the information is not present in the treebank.
On the contrary, in this work, we directly score parts of the syntactic content (i.e.\ labeled constituents).
Therefore, at training time we can optimize an objective directly related to the end-goal evaluation.

{\bf Transition systems:}
Lastly, transition-based parsers have been proposed, based on the idea of the \textsc{Swap} transition for non-projective dependency parsing \cite{nivre2009swap}, see \newcite{versley2014swapconst} and following work based on shift-reduce strategy \cite{maier2015shiftreduce,maier2016discontinuous,stanojevic2017neuralshift}.
These systems rely on the fact that a discontinuous tree can be transformed into a continuous one by changing word order in the input sentence.
They do not require strong independence assumption on the scoring model which can be useful to encode richer information, especially for long-distance relationships.
However, the number of transitions required to parse discontinuities can impact prediction accuracy.
To alleviate this problem, two different approaches have been explored:
\newcite{coavoux2017incremental} introduced a two stacks system coupled with a \textsc{Gap} transition and \newcite{maier2016discontinuous} proposed the \textsc{Shift-i} transition to access non-local elements directly, therefore reducing the number of transitions.
In exchange for a rich parameterization, transition systems lose optimality guarantees with respect to the scoring model and rely on greedy or beam-search decoding that can return sub-optimal solutions.
These approaches achieve state-of-the-art results while being fast at test time \cite{coavoux2019stackfree,coavoux2019unlex}.
On the contrary, our approach is exact with respect to the scoring model, i.e.\ it will always return the highest scoring structure.

\section{Parsing Algorithm}
\label{sec:alg}

We describe our algorithm using the \emph{parsing-as-deduction} framework \cite{pereira1983deduction}.
As such, our description is independent of the value one wants to compute, whether it be the (k-)best derivation(s), partition function or span marginals \cite{goodman1999semiring}.%
\footnote{Note that parsing without grammatical constraints results in all sentences having a non-empty parse forest, therefore the recognition problem is ill-defined.}
However, we will focus on argmax decoding.

We are interested in constituency trees of block degree two, including ill-nested trees.
The block degree two constraint is satisfied if each node dominate at most two disjoint sequences of words.
Let $w_1 ... w_n$ be a sentence.
A constituency tree over this sentence is ill-nested if it contains two nodes dominating disjoint sets of words $W^{(1)}$ and $W^{(2)}$ such that there exists $w_i, w_j \in W^{(1)}$ and $w_k, w_l \in W^{(2)}$ such that $i < k < j < l$ or $k < i < l < j$.

We first highlight some properties of span-based parsers:
\begin{itemize}
	\item {\bf Filtering:}
		Contrary to CFGs and LCFRS CKY-style parsers, there is no side-condition constraining allowed derivations in span-based parsers.
		The label of a constituent is independent of the label of its children.
	\item {\bf Binarization:}
		Interestingly, span-based parsers do not require explicit binarization of the constituency structure.
		Although grammar based parsers require binarization of the grammar production rules and therefore of the constituency structure to ensure tractable complexity,
 span-based parsers can take care of binarization \emph{implicitly} by introducing a supplementary constituency label with a fixed null score.
	\item {\bf Unary rules:}
		We follow \newcite{stern2017spanbased} and merge unary chains into a single constituent with a new label, e.g. the chain $\texttt{SBARQ} \rightarrow \texttt{SQ}$ will result in a single constituent labeled $\texttt{SBARQ\_SQ}$.
\end{itemize}

\subsection{Items}

Let $\mathcal{N}$ be the set of non-terminals (labels) and $n$ the length of the input sentence.
We define spans with interstice indices instead of word indices, see Figure~\ref{fig:example}.
Items manipulated by our deduction rules are 5-tuples $[A, i, k, l, j]$ where $A \in \mathcal{N} \cup \{ \varnothing \}$ is the constituent label with value $\varnothing$ indicating an empty span used for implicit binarization.
Given that each item represent a constituent, we will use the same notation to refer to the chart item and to the linguistic structure interchangeably.
Indices $i, j \in \mathbb{N}, k, l \in \mathbb{N} \cup \{ - \}$ defines the constituent span as follows:
\begin{itemize}
    \item if the constituent is continuous, then $k = l = -$ and $0 \leq i < j \leq n$;
    \item
    	otherwise, the constituent is discontinuous (with a single gap) and $0 \leq i < k$ and $l < j \leq n$, with $k < l$, define its left and right spans, respectively.
    	For example, the tree in Figure~\ref{fig:example} contains the discontinuous constituent $[VP, 0, 1, 5, 6]$.
\end{itemize}

\subsection{Axioms and goal}

Axiom items are word level constituents, i.e.\ items of the form $[A, i, -, -, i + 1]$
with $0 \leq i < n$ and $A \in \mathcal{N} \cup \{ \varnothing \}$.
In our experiments, axioms can have a null label, i.e.\ $A = \varnothing$, because we do not include part-of-speech tags as leaves of the constituency tree.

The goal item is defined as $[A, 0, -, -, n]$
with  $A \in \mathcal{N} \cup \{ \varnothing \}$.
Similarly, in our experiments, the goal can have a null label, so we can parse empty trees and disconnected structures present in treebanks without further pre/post-processing steps.

\subsection{Deduction rules}

\begin{figure*}[t]
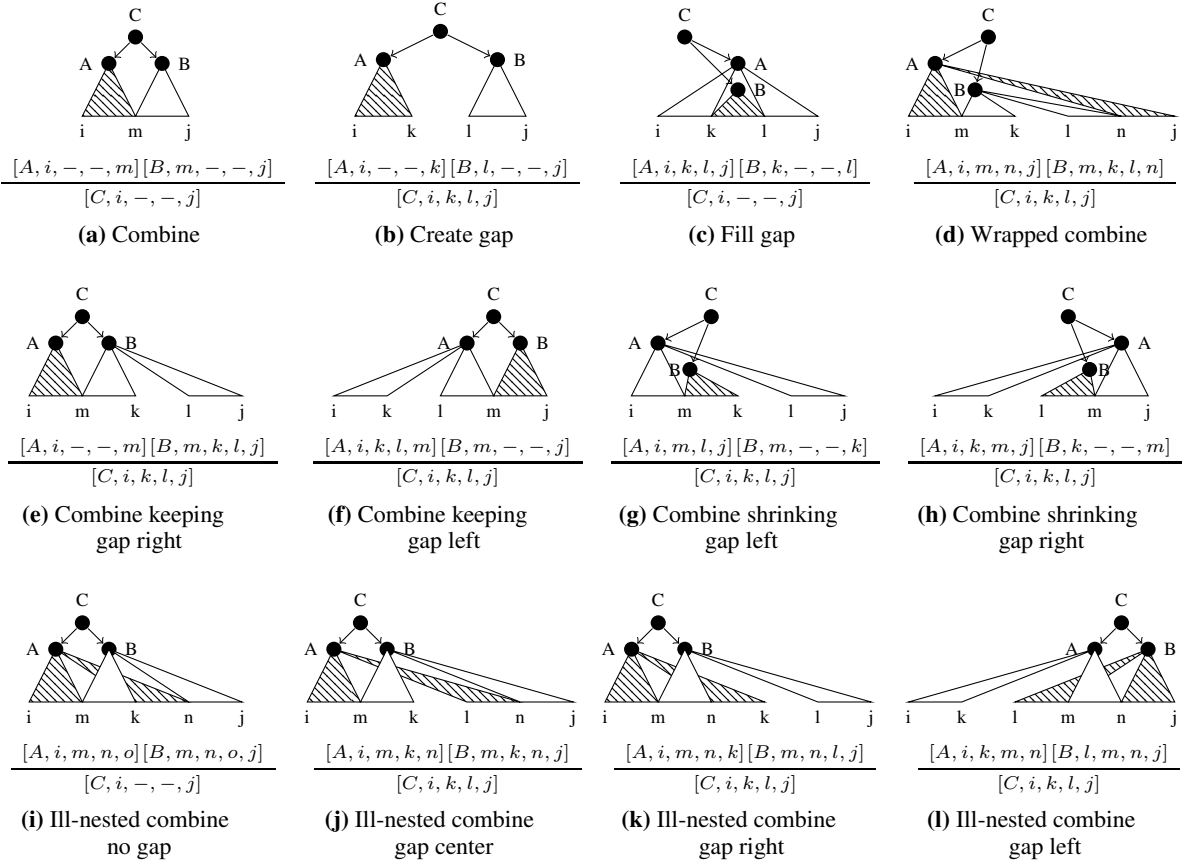


\begin{tabular}{@{}c@{}@{}c@{}@{}c@{}@{}c@{}}
		\begin{tikzpicture}[scale=0.7, cst/.style={circle,fill=black,inner sep=2pt}, every label/.style={label distance=0.1cm,inner sep=0pt}]
	\node[label={[label distance=-0.2cm]below:\strut\scriptsize i}] (i) at (0, 0) {};
	\node[label={[label distance=-0.2cm]below:\strut\scriptsize m}] (m) at (1,0) {};
	\node[label={[label distance=-0.2cm]below:\strut\scriptsize j}] (j) at (2,0) {};
	
	\node[cst,label=left:{\scriptsize A}] (t1) at (0.5,1) {};
	\node[cst,label=right:{\scriptsize B}] (t2) at (1.5,1) {};
	\node[cst,label=above:{\scriptsize C}] (t3) at (1,1.5) {};
	
	\draw[->] (t3) -- (t1);
	\draw[->] (t3) -- (t2);
	
	\draw[pattern=north west lines] (t1.center) -- (i.center) -- (m.center) -- (t1.center);
	\draw (t2.center) -- (m.center) -- (j.center) -- (t2.center);
	\end{tikzpicture}
	& \begin{tikzpicture}[scale=0.75, cst/.style={circle,fill=black,inner sep=2pt}, every label/.style={label distance=0.1cm,inner sep=0pt}]
\node[label={[label distance=-0.2cm]below:\strut\scriptsize i}] (i) at (0, 0) {};
\node[label={[label distance=-0.2cm]below:\strut\scriptsize k}] (k) at (1,0) {};
\node[label={[label distance=-0.2cm]below:\strut\scriptsize l}] (l) at (2, 0) {};
\node[label={[label distance=-0.2cm]below:\strut\scriptsize j}] (j) at (3,0) {};

\node[cst,label=left:{\scriptsize A}] (t1) at (0.5,1) {};
\node[cst,label=right:{\scriptsize B}] (t2) at (2.5,1) {};
\node[cst,label=above:{\scriptsize C}] (t3) at (1.5,1.5) {};

\draw[->] (t3) -- (t1);
\draw[->] (t3) -- (t2);

\draw[pattern=north west lines] (t1.center) -- (i.center) -- (k.center) -- (t1.center);
\draw (t2.center) -- (l.center) -- (j.center) -- (t2.center);
\end{tikzpicture}
	& 	\begin{tikzpicture}[scale=0.7, cst/.style={circle,fill=black,inner sep=2pt}, every label/.style={label distance=0.1cm,inner sep=0pt}]
\node[label={[label distance=-0.2cm]below:\strut\scriptsize i}] (i) at (0, 0) {};
\node[label={[label distance=-0.2cm]below:\strut\scriptsize k}] (k) at (1,0) {};
\node[label={[label distance=-0.2cm]below:\strut\scriptsize l}] (l) at (2, 0) {};
\node[label={[label distance=-0.2cm]below:\strut\scriptsize j}] (j) at (3,0) {};

\node[cst,label=right:{\scriptsize A}] (t1) at (1.5,1) {};
\node[cst,label=right:{\scriptsize B}] (t2) at (1.5,0.5) {};
\node[cst,label=above:{\scriptsize C}] (t3) at (0.5,1.5) {};

\draw[->] (t3) -- (t1);
\draw[->] (t3) -- (t2);

\draw[pattern=north west lines] (t2.center) -- (k.center) -- (l.center) -- (t2.center);
\draw (t1.center) -- (i.center) -- (k.center) -- (t1.center) -- (l.center) -- (j.center) -- (t1.center);
\end{tikzpicture}
	& 	\begin{tikzpicture}[scale=0.7, cst/.style={circle,fill=black,inner sep=2pt}, every label/.style={label distance=0.1cm,inner sep=0pt}]
	\node[label={[label distance=-0.2cm]below:\strut\scriptsize i}] (i) at (0, 0) {};
	\node[label={[label distance=-0.2cm]below:\strut\scriptsize m}] (m) at (1,0) {};
	\node[label={[label distance=-0.2cm]below:\strut\scriptsize k}] (k) at (2,0) {};
	\node[label={[label distance=-0.2cm]below:\strut\scriptsize l}] (l) at (3,0) {};
	\node[label={[label distance=-0.2cm]below:\strut\scriptsize n}] (n) at (4,0) {};
	\node[label={[label distance=-0.2cm]below:\strut\scriptsize j}] (j) at (5,0) {};
	
	\node[cst,label=left:{\scriptsize A}] (t1) at (0.5,1) {};
	\node[cst,label={[label distance=0cm]left:\scriptsize B}] (t2) at (1.25,0.5) {};
	\node[cst,label=above:{\scriptsize C}] (t3) at (1.5,1.5) {};
	
	\draw[->] (t3) -- (t1);
	\draw[->] (t3) -- (t2);
	
	\draw[pattern=north west lines] (t1.center) -- (i.center) -- (m.center) -- (t1.center) -- (n.center) -- (j.center) -- (t1.center);
	\draw (t2.center) -- (m.center) -- (k.center) -- (t2.center) -- (l.center) -- (n.center) -- cycle;
	\end{tikzpicture}
	\\
	\subcaptionbox{\centering Combine\label{rule:n3}}{\input{deductions/combine.rule.tex}}
	& \subcaptionbox{\centering Create gap\label{rule:create_gap}}{\input{deductions/create_gap.rule.tex}}
	& \subcaptionbox{\centering Fill gap\label{rule:fill_gap}}{\input{deductions/fill_gap.rule.tex}}
	& \subcaptionbox{\centering Wrapped combine\label{rule:wrapped}}{\input{deductions/wrap.rule.tex}}
	\vspace{0.5cm}\\
		\begin{tikzpicture}[scale=0.7, cst/.style={circle,fill=black,inner sep=2pt}, every label/.style={label distance=0.1cm,inner sep=0pt}]
\node[label={[label distance=-0.2cm]below:\strut\scriptsize i}] (i) at (0, 0) {};
\node[label={[label distance=-0.2cm]below:\strut\scriptsize m}] (m) at (1,0) {};

\node[cst,label=left:{\scriptsize A}] (t1) at (0.5,1) {};
\node[cst,label=right:{\scriptsize B}] (t2) at (1.5,1) {};
\node[cst,label=above:{\scriptsize C}] (t3) at (1,1.5) {};

\draw[->] (t3) -- (t1);
\draw[->] (t3) -- (t2);

\node[label={[label distance=-0.2cm]below:\strut\scriptsize k}] (k) at (2,0) {};
\node[label={[label distance=-0.2cm]below:\strut\scriptsize l}] (l) at (3,0) {};
\node[label={[label distance=-0.2cm]below:\strut\scriptsize j}] (j) at (4,0) {};

\draw[pattern=north west lines] (t1.center) -- (i.center) -- (m.center) -- (t1.center);
\draw (t2.center) -- (j.center) -- (l.center) -- (t2.center) -- (k.center) --(m.center) -- (t2.center);
\end{tikzpicture}
	& 	\begin{tikzpicture}[scale=0.7, cst/.style={circle,fill=black,inner sep=2pt}, every label/.style={label distance=0.1cm,inner sep=0pt}]
\node[label={[label distance=-0.2cm]below:\strut\scriptsize i}] (i) at (0, 0) {};
\node[label={[label distance=-0.2cm]below:\strut\scriptsize k}] (k) at (1,0) {};
\node[label={[label distance=-0.2cm]below:\strut\scriptsize l}] (l) at (2,0) {};
\node[label={[label distance=-0.2cm]below:\strut\scriptsize m}] (m) at (3,0) {};
\node[label={[label distance=-0.2cm]below:\strut\scriptsize j}] (j) at (4,0) {};

\node[cst,label=right:{\scriptsize B}] (t1) at (3.5,1) {};
\node[cst,label=left:{\scriptsize A}] (t2) at (2.5,1) {};
\node[cst,label=above:{\scriptsize C}] (t3) at (3,1.5) {};

\draw[->] (t3) -- (t1);
\draw[->] (t3) -- (t2);

\draw[pattern=north west lines] (t1.center) -- (m.center) -- (j.center) -- (t1.center);
\draw (t2.center) -- (i.center) -- (k.center) -- (t2.center) -- (l.center) --(m.center) -- (t2.center);
\end{tikzpicture}
	& 	\begin{tikzpicture}[scale=0.7, cst/.style={circle,fill=black,inner sep=2pt}, every label/.style={label distance=0.1cm,inner sep=0pt}]
\node[label={[label distance=-0.2cm]below:\strut\scriptsize i}] (i) at (0, 0) {};
\node[label={[label distance=-0.2cm]below:\strut\scriptsize m}] (m) at (1,0) {};
\node[label={[label distance=-0.2cm]below:\strut\scriptsize k}] (k) at (2,0) {};
\node[label={[label distance=-0.2cm]below:\strut\scriptsize l}] (l) at (3,0) {};
\node[label={[label distance=-0.2cm]below:\strut\scriptsize j}] (j) at (4,0) {};

\node[cst,label=left:{\scriptsize A}] (t1) at (0.5,1) {};
\node[cst,label={[label distance=0cm]left:\scriptsize B}] (t2) at (1.1,0.5) {};
\node[cst,label=above:{\scriptsize C}] (t3) at (1.5,1.5) {};

\draw[->] (t3) -- (t1);
\draw[->] (t3) -- (t2);

\draw[pattern=north west lines] (t2.center) -- (m.center) -- (k.center) -- (t2.center);
\draw (t1.center) -- (i.center) -- (m.center) -- (t1.center) -- (l.center) --(j.center) -- (t1.center);
\end{tikzpicture}
	& 	\begin{tikzpicture}[scale=0.7, cst/.style={circle,fill=black,inner sep=2pt}, every label/.style={label distance=0.1cm,inner sep=0pt}]
\node[label={[label distance=-0.2cm]below:\strut\scriptsize i}] (i) at (0, 0) {};
\node[label={[label distance=-0.2cm]below:\strut\scriptsize k}] (k) at (1,0) {};
\node[label={[label distance=-0.2cm]below:\strut\scriptsize l}] (l) at (2,0) {};
\node[label={[label distance=-0.2cm]below:\strut\scriptsize m}] (m) at (3,0) {};
\node[label={[label distance=-0.2cm]below:\strut\scriptsize j}] (j) at (4,0) {};

\node[cst,label=right:{\scriptsize A}] (t1) at (3.5,1) {};
\node[cst,label={[label distance=0cm]right:\scriptsize B}] (t2) at (2.9,0.5) {};
\node[cst,label=above:{\scriptsize C}] (t3) at (2.5,1.5) {};

\draw[->] (t3) -- (t1);
\draw[->] (t3) -- (t2);

\draw[pattern=north west lines] (t2.center) -- (l.center) -- (m.center) -- (t2.center);
\draw (t1.center) -- (i.center) -- (k.center) -- (t1.center) -- (m.center) --(j.center) -- (t1.center);
\end{tikzpicture}
	\\
	\subcaptionbox{\centering Combine keeping\newline gap right\label{rule:keep_right}}{\input{deductions/combine_ol.rule.tex}}
	& \subcaptionbox{\centering Combine keeping\newline gap left\label{rule:keep_left}}{\input{deductions/combine_or.rule.tex}}
	& \subcaptionbox{\centering Combine shrinking\newline gap left\label{rule:shrink_left}}{\input{deductions/combine_il.rule.tex}}
	& \subcaptionbox{\centering Combine shrinking\newline gap right\label{rule:shrink_right}}{\input{deductions/combine_ir.rule.tex}}
	\vspace{0.5cm}\\
		\begin{tikzpicture}[scale=0.7, cst/.style={circle,fill=black,inner sep=2pt}, every label/.style={label distance=0.1cm,inner sep=0pt}]
\node[label={[label distance=-0.2cm]below:\strut\scriptsize i}] (i) at (0, 0) {};
\node[label={[label distance=-0.2cm]below:\strut\scriptsize m}] (m) at (1,0) {};
\node[label={[label distance=-0.2cm]below:\strut\scriptsize k}] (k) at (2,0) {};
\node[label={[label distance=-0.2cm]below:\strut\scriptsize n}] (n) at (3,0) {};
\node[label={[label distance=-0.2cm]below:\strut\scriptsize j}] (j) at (4,0) {};

\node[cst,label=left:{\scriptsize A}] (t1) at (0.5,1) {};
\node[cst,label=right:{\scriptsize B}] (t2) at (1.5,1) {};
\node[cst,label=above:{\scriptsize C}] (t3) at (1,1.5) {};

\draw[->] (t3) -- (t1);
\draw[->] (t3) -- (t2);

\draw[pattern=north west lines] (t1.center) -- (i.center) -- (m.center) -- (t1.center) -- (k.center) -- (n.center) -- (t1.center);
\draw[fill=white] (t2.center) -- (m.center) -- (k.center) -- (t2.center) -- (n.center) -- (j.center) -- cycle;
\end{tikzpicture}
	& 	\begin{tikzpicture}[scale=0.7, cst/.style={circle,fill=black,inner sep=2pt}, every label/.style={label distance=0.1cm,inner sep=0pt}]
\node[label={[label distance=-0.2cm]below:\strut\scriptsize i}] (i) at (0, 0) {};
\node[label={[label distance=-0.2cm]below:\strut\scriptsize m}] (m) at (1,0) {};
\node[label={[label distance=-0.2cm]below:\strut\scriptsize k}] (k) at (2,0) {};
\node[label={[label distance=-0.2cm]below:\strut\scriptsize l}] (l) at (3,0) {};
\node[label={[label distance=-0.2cm]below:\strut\scriptsize n}] (n) at (4,0) {};
\node[label={[label distance=-0.2cm]below:\strut\scriptsize j}] (j) at (5,0) {};

\node[cst,label=left:{\scriptsize A}] (t1) at (0.5,1) {};
\node[cst,label=right:{\scriptsize B}] (t2) at (1.5,1) {};
\node[cst,label=above:{\scriptsize C}] (t3) at (1,1.5) {};

\draw[->] (t3) -- (t1);
\draw[->] (t3) -- (t2);

\draw[pattern=north west lines] (t1.center) -- (i.center) -- (m.center) -- (t1.center) -- (l.center) -- (n.center) -- (t1.center);

\draw[fill=white] (t2.center) -- (m.center) -- (k.center) -- (t2.center) -- (n.center) -- (j.center) -- cycle;
\end{tikzpicture}
	& 	\begin{tikzpicture}[scale=0.7, cst/.style={circle,fill=black,inner sep=2pt}, every label/.style={label distance=0.1cm,inner sep=0pt}]
\node[label={[label distance=-0.2cm]below:\strut\scriptsize i}] (i) at (0, 0) {};
\node[label={[label distance=-0.2cm]below:\strut\scriptsize m}] (m) at (1,0) {};
\node[label={[label distance=-0.2cm]below:\strut\scriptsize n}] (n) at (2,0) {};
\node[label={[label distance=-0.2cm]below:\strut\scriptsize k}] (k) at (3,0) {};
\node[label={[label distance=-0.2cm]below:\strut\scriptsize l}] (l) at (4,0) {};
\node[label={[label distance=-0.2cm]below:\strut\scriptsize j}] (j) at (5,0) {};

\node[cst,label=left:{\scriptsize A}] (t1) at (0.5,1) {};
\node[cst,label=right:{\scriptsize B}] (t2) at (1.5,1) {};
\node[cst,label=above:{\scriptsize C}] (t3) at (1,1.5) {};

\draw[->] (t3) -- (t1);
\draw[->] (t3) -- (t2);

\draw[pattern=north west lines] (t1.center) -- (i.center) -- (m.center) -- (t1.center) -- (n.center) -- (k.center) -- (t1.center);

\draw[fill=white] (t2.center) -- (m.center) -- (n.center) -- (t2.center) -- (l.center) -- (j.center) -- cycle;
\end{tikzpicture}
	& 	\begin{tikzpicture}[scale=0.7, cst/.style={circle,fill=black,inner sep=2pt}, every label/.style={label distance=0.1cm,inner sep=0pt}]
\node[label={[label distance=-0.2cm]below:\strut\scriptsize i}] (i) at (0, 0) {};
\node[label={[label distance=-0.2cm]below:\strut\scriptsize k}] (k) at (1,0) {};
\node[label={[label distance=-0.2cm]below:\strut\scriptsize l}] (l) at (2,0) {};
\node[label={[label distance=-0.2cm]below:\strut\scriptsize m}] (m) at (3,0) {};
\node[label={[label distance=-0.2cm]below:\strut\scriptsize n}] (n) at (4,0) {};
\node[label={[label distance=-0.2cm]below:\strut\scriptsize j}] (j) at (5,0) {};

\node[cst,label=right:{\scriptsize B}] (t1) at (4.5,1) {};
\node[cst,label=left:{\scriptsize A}] (t2) at (3.5,1) {};
\node[cst,label=above:{\scriptsize C}] (t3) at (4,1.5) {};

\draw[->] (t3) -- (t1);
\draw[->] (t3) -- (t2);

\draw[pattern=north west lines] (t1.center) -- (l.center) -- (m.center) -- (t1.center) -- (n.center) -- (j.center) -- (t1.center);
\draw[fill=white] (t2.center) -- (i.center) -- (k.center) -- (t2.center) -- (m.center) -- (n.center) -- cycle;

\end{tikzpicture}
	\\
	\subcaptionbox{\centering Ill-nested combine\newline no gap\label{rule:ill1}}{\input{deductions/il_no_gap.rule.tex}}
	& \subcaptionbox{\centering Ill-nested combine\newline  gap center\label{rule:ill2}}{\input{deductions/il_gap.rule.tex}}
	& \subcaptionbox{\centering Ill-nested combine\newline  gap right\label{rule:ill3}}{\input{deductions/il_gap_right.rule.tex}}
	& \subcaptionbox{\centering Ill-nested combine\newline  gap left\label{rule:ill4}}{\input{deductions/il_gap_left.rule.tex}}
\end{tabular}

    \caption{Deduction rules of our algorithm.}
    \label{fig:rules}
\end{figure*}

The deduction rules used to derive the goal from axioms are listed on Figure~\ref{fig:rules}.
Each rule takes exactly two antecedents.
Note that rule \ruleref{rule:n3} is the single rule used by span-based continuous constituency parsers.

Rule \ruleref{rule:create_gap} creates a discontinuous constituent from two continuous constituents.
The set of rules \ruleref{rule:keep_right}-\ruleref{rule:keep_left}-\ruleref{rule:shrink_left}-\ruleref{rule:shrink_right} (resp.\ \ruleref{rule:fill_gap}) allow to combine one discontinuous and one continuous constituent to produce a discontinuous one (resp.\ a continuous one).

Finally, there are rules that combine two discontinuous antecedents.
Rule \ruleref{rule:wrapped} is the only such rule that is allowed for building well-nested trees.
The other four rules \ruleref{rule:ill1}-\ruleref{rule:ill2}-\ruleref{rule:ill3}-\ruleref{rule:ill4} are used for the construction of ill-nested trees.
As such, it is easy to control whether ill-nested structures are permitted or not by including or excluding them.

\subsection{Soundness and completness}

On the one hand, the algorithm is sound by definition because:
\begin{itemize}
	\item items cannot represent constituents with a gap degree strictly greater to two;
	\item every rule deduces an item representing a constituent spanning a greater number of words, therefore they cannot construct invalid trees where the parent of constituent spans fewer words than one of its children.
\end{itemize}
On the other hand, completeness can be proved by observing that every possible binary parent-children combination can be produced by one of the rule.
For the non-binary case, completeness follows the fact the fact a constituent with 3 or more children can be built by first deriving intermediary constituents with label $\varnothing$.

Note that due to implicit binarization, a non-binary tree can be constructed by different sequences of deductions.
Therefore, special care must be taken for computing partition function and marginal probabilities.
As we are not interested by these values in this work, we do not dwell into this issue.

\subsection{Complexity}

The space and time complexity can be inferred from item structures and deduction rules: the space complexity is $\mathcal{O}(|\mathcal{N}|n^4)$ and time complexity is $\mathcal{O}(|\mathcal{N}|^3n^6)$.
In practice, we decompose the score of a tree into the sum of the score of its constituents only and there are no constraints between antecedents and consequent labels.
Therefore, we can build intermediary unlabeled items as follow:
$$
\AxiomC{$[A, i, k, l, j]$}
\UnaryInfC{$[i, k, l, j]$}
\DisplayProof
$$
which replace antecedents in every rule in Figure~\ref{fig:rules}.
With this update, the time complexity is linear in the number of labels, that is, $\mathcal{O}(|\mathcal{N}|n^6)$.

We instantiate variants of the algorithm than cannot parse the full family of block degree two trees but that can still fit most actual linguistic structures present in treebanks, with a lower time complexity.
By using only rules \ruleref{rule:n3}, \ruleref{rule:create_gap} and \ruleref{rule:fill_gap} we can build a parser with a $\mathcal{O}(n^4)$ time complexity.
In the next section, we show that this specific variant can be optimized into a  $\mathcal{O}(n^3)$ time parser.
By adding rules \ruleref{rule:keep_right}, \ruleref{rule:keep_left}, \ruleref{rule:shrink_left}, \ruleref{rule:shrink_right} and \ruleref{rule:ill1} we build a $\mathcal{O}(n^5)$ parser.
Finally, we construct $\mathcal{O}(n^5)$ and $\mathcal{O}(n^6)$ well-nested parsers by excluding rules \ruleref{rule:ill1}, \ruleref{rule:ill2}, \ruleref{rule:ill3} and \ruleref{rule:ill4}.

\subsection{Cubic time discontinuous constituency parser}

A specific variant uses only deduction rules \ruleref{rule:n3}, \ruleref{rule:create_gap} and \ruleref{rule:fill_gap} from Figure~\ref{fig:rules}, leading to a $\mathcal O(n^4)$ space and time complexity.
In this setting, there is no way to combine two items representing discontinuous constituents or to have a discontinuous constituent that has a discontinuous child in the resulting parse tree.
In this section, we prove that the family of trees induced by this variant of the parser can actually be parsed with a $\mathcal O(n^3)$ time complexity, that is equivalent to continuous constituency parsers.

\begin{figure*}[t]
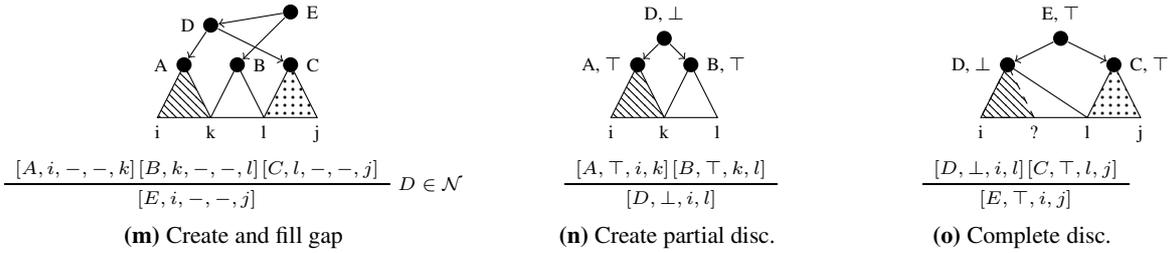

	\centering
	\foreach \n in {1,...,12}{\phantomsubcaption}
	\begin{tabular}{ccc}
		\hfill\begin{tikzpicture}[scale=0.7, cst/.style={circle,fill=black,inner sep=2pt}, every label/.style={label distance=0.1cm,inner sep=0pt}]

\node[label={[label distance=-0.2cm]below:\strut\scriptsize i}] (i) at (0, 0) {};
\node[label={[label distance=-0.2cm]below:\strut\scriptsize k}] (k) at (1,0) {};
\node[label={[label distance=-0.2cm]below:\strut\scriptsize l}] (l) at (2,0) {};
\node[label={[label distance=-0.2cm]below:\strut\scriptsize j}] (j) at (3,0) {};

\node[cst,label=left:{\scriptsize A}] (t1) at (0.5,1) {};
\node[cst,label=right:{\scriptsize B}] (t2) at (1.5,1) {};
\node[cst,label=right:{\scriptsize C}] (t3) at (2.5,1) {};
\node[cst,label=left:{\scriptsize D}] (t4) at (1,1.75) {};
\node[cst,label=right:{\scriptsize E}] (t5) at (2.5,2) {};

\draw[->] (t4) -- (t1);
\draw[->] (t4) -- (t3);
\draw[->] (t5) -- (t2);
\draw[->] (t5) -- (t4);

\draw[pattern=north west lines] (t1.center) -- (i.center) -- (k.center) -- (t1.center);
\draw (t2.center) -- (k.center) -- (l.center) -- (t2.center);
\draw[pattern=dots] (t3.center) -- (l.center) -- (j.center) -- (t3.center);
\end{tikzpicture}
		\hspace{1cm}
		&
		\hspace{1cm}
		\begin{tikzpicture}[scale=0.7, cst/.style={circle,fill=black,inner sep=2pt}, every label/.style={label distance=0.1cm,inner sep=0pt}]
\node[label={[label distance=-0.2cm]below:\strut\scriptsize i}] (i) at (0, 0) {};
\node[label={[label distance=-0.2cm]below:\strut\scriptsize k}] (k) at (1,0) {};
\node[label={[label distance=-0.2cm]below:\strut\scriptsize l}] (l) at (2, 0) {};

\node[cst,label=left:{\scriptsize A, $\top$}] (t1) at (0.5,1) {};
\node[cst,label=right:{\scriptsize B, $\top$}] (t2) at (1.5,1) {};
\node[cst,label=above:{\scriptsize D, $\bot$}] (t3) at (1,1.5) {};

\draw[->] (t3) -- (t1);
\draw[->] (t3) -- (t2);

\draw[pattern=north west lines] (t1.center) -- (i.center) -- (k.center) -- (t1.center);
\draw (t2.center) -- (k.center) -- (l.center) -- (t2.center);
\end{tikzpicture}
		\hspace{1cm}
		&
		\hspace{1cm}
		\begin{tikzpicture}[scale=0.7, cst/.style={circle,fill=black,inner sep=2pt}, every label/.style={label distance=0.1cm,inner sep=0pt}]
\node[label={[label distance=-0.2cm]below:\strut\scriptsize i}] (i) at (0, 0) {};
\node[label={[label distance=-0.2cm]below:\strut\scriptsize ?}] (k) at (1,0) {};
\node[label={[label distance=-0.2cm]below:\strut\scriptsize l}] (l) at (2, 0) {};
\node[label={[label distance=-0.2cm]below:\strut\scriptsize j}] (j) at (3, 0) {};

\node[cst,label=left:{\scriptsize D, $\bot$}] (t1) at (0.5,1) {};
\node[cst,label=right:{\scriptsize C, $\top$}] (t2) at (2.5,1) {};
\node[cst,label=above:{\scriptsize E, $\top$}] (t3) at (1.5,1.5) {};

\draw[->] (t3) -- (t1);
\draw[->] (t3) -- (t2);

\draw[pattern=north west lines,solid] (t1.center) -- (i.center) -- (k.center)  [dashed]-- (t1.center);
\draw[solid] (t1.center) -- (i.center) -- (k.center);
\draw (t1.center) -- (l.center) -- (k.center);
\draw[pattern=dots] (t2.center) -- (l.center) -- (j.center) -- (t2.center);
\end{tikzpicture}
		\\
		\subcaptionbox{\centering Create and fill gap\label{rule:create_and_combine}}{\input{deductions/create_and_fill.rule.tex}}
		& \subcaptionbox{\centering Create partial disc.\label{rule:n3_1}}{\input{deductions/n3_combine_gap1.rule.tex}}
		& \subcaptionbox{\centering Complete disc.\label{rule:n3_2}}{\input{deductions/n3_combine_gap2.rule.tex}}
	\end{tabular}
\caption{%
	\ruleref{rule:create_and_combine} The create gap and fill gap rules can be merged into a single rule if there are no other rule with discontinuous antecedents in the parser.
	\ruleref{rule:n3_1}-\ruleref{rule:n3_2} Rules for the cubic time discontinuous constituency parser. The rule to combine two continuous constituents follows the previous one.
}
\label{fig:cubic_rules}

\end{figure*}

The intuition goes as follows. We could replace rules \ruleref{rule:create_gap} and \ruleref{rule:fill_gap} with the single rule \ruleref{rule:create_and_combine} in Figure~\ref{fig:cubic_rules}
where the right hand side condition $D \in \mathcal N$ induces the existence of a discontinuous constituent with label $D$,%
\footnote{Without loss of generality, we assume the label $D$ is not null. Although it could be without changing the overall idea, we would just add an extra way to do implicit binarization that can already be handled with rule \ruleref{rule:n3}.} %
with is left part spanning words $i$ to $k$ and right part spanning words $l$ to $j$.
However, observe that this new rule realizes two tests that could be done independently:
\begin{enumerate}
	\item the right span boundary of the first antecedent must match the left span boundary of the second one;
	\item  the right span boundary of the second antecedent must match the left span boundary of the third antecedent.
\end{enumerate}
Therefore, we can break the deduction into two sequential deductions, first testing the "$k$" boundary then the "$l"$ boundary.%
\footnote{This idea of breaking up simultaneous tests in a deduction rule has been previously proposed for improving time complexity of lexicalized grammar parsers \cite{eisner1999bilexical,eisner2000fasterltag}}

To this end, we build a parser based on 4-tuple items $[A, \tau, i, j]$ where $\tau \in \{\top, \bot\}$ indicates whether the item represents a continuous constituent ($\tau = \top$) or an incomplete discontinuous constituent ($\tau = \bot$).
More precisely, an item $[A, \bot, i, j]$ represents a partial discontinuous constituent who would be represented as $[A, i, ?, j, ?]$ in the previous formalization.
The right boundary of its two spans are unknown: the one of the left span has been "forgotten" and the one on the right span is yet to be determined.
The deduction rules of this new parser are listed on Figure~\ref{fig:cubic_rules}, with axioms $[A, \tau, i, i+1]$, $0 \leq i < n$, and goal $[A, \tau, 0, n]$.

Note that this cubic time algorithm imposes an additional restriction for weighted parsing:
the score of discontinuous constituent must be divided into smaller sub-parts, which we do in practice for all deduction systems due to computational reasons.

We report the running time per sentence length for the $\mathcal{O}(n^4)$ and $\mathcal{O}(n^3)$ parsers in Figure~\ref{fig:optimization}.
As expected, the running time of the cubic time parser is way lower for long sentence.

\section{Experiments}

We experiment on the Discontinuous Penn Treebank \cite[\dptb{}]{marcus1993ptb,evang2011plcfrs} with standard split, the \tiger{} treebank \cite{brants2002tiger} with the SPMRL 2014 shared task split \cite{seddah2014spmrl} and the \negra{} treebank \cite{skut1997negra} with the split proposed by \newcite{dubey2003negra}.

\subsection{Data coverage}

One important question is whether our parser has a good coverage of the dataset as we can only retrieve constituents of block degree one and two.
We report the maximum recall that our parser can achieve in its different variants in Table~\ref{table:cover}.

First, we observe that our cubic time parser can recover 98\% of all constituents in the three treebanks, or around 80\% of constituents of block degree of exactly two.
Second, the $\mathcal O(n^5)$ variant of the parser can recover more than 99\% of all treebanks, and, interestingly, there is almost no coverage change when moving to the full deduction system. 
If we consider the parsers with well-nested restriction, the $\mathcal O(n^5)$ and $\mathcal O(n^6)$ variants have the same coverage in German datasets and the later can only recover 2 additional constituents in the English treebanks.
If we include ill-nested construction, the difference is either 2 (\dptb{} and \negra{}) or 8 (\tiger{}) constituents.
In practice, we observed that both $\mathcal O(n^5)$ and $\mathcal O(n^6)$ variants predict the same analysis.

\begin{table*}[p]
	\centering
	\small
	\begin{tabular}{@{}crcccccc@{}}
		\toprule
		&& Continuous &  $\mathcal O(n^3)$ &  $\mathcal O(n^5)$ / WN &  $\mathcal O(n^5)$ &  $\mathcal O(n^6)$ / WN &  $\mathcal O(n^6)$ \\
		\hline
		\parbox[t]{2mm}{\multirow{4}{*}{\rotatebox[origin=c]{90}{DPTB}}}
		& All & 98.16 & 99.46 & 99.81 & 99.83 & 99.81 & 99.83 \\
		& BD $\leq 2$ & 98.32 & 99.63 & 99.98 & 99.99 & 99.98 & 100.00 \\
		& BD $= 2$ & 0.00 & 78.27 & 99.15 & 99.98 & 99.17 & 100.00 \\
		& & (0) & (10713) & (13572) & (13685) & (13574) & (13687) \\
		\midrule
		\parbox[t]{2mm}{\multirow{4}{*}{\rotatebox[origin=c]{90}{TIGER}}}
		& All & 94.51 & 98.61 & 99.37 & 99.49 & 99.37 & 99.49 \\
		& BD $\leq 2$ & 94.99 & 99.11 & 99.88 & 99.99 & 99.88 & 100.00 \\
		& BD $= 2$ & 0.00 & 82.39 & 97.65 & 99.95 & 97.65 & 100.00 \\
		& & (0) & (15324) & (18161) & (18590) & (18161) & (18598) \\
		\midrule
		\parbox[t]{2mm}{\multirow{4}{*}{\rotatebox[origin=c]{90}{NEGRA}}}
		&All & 94.37 & 98.59 & 99.32 & 99.46 & 99.32 & 99.46 \\
		& BD $\leq 2$ & 94.87 & 99.12 & 99.85 & 99.99 & 99.85 & 100.00 \\
		& BD $= 2$ & 0.00 & 82.91 & 97.24 & 99.97 & 97.24 & 100.00 \\
		& & (0) & (6106) & (7161) & (7362) & (7161) & (7364) \\
		\bottomrule
	\end{tabular}
	\caption{%
		Maximum constituent recall that be can obtained using a continuous constituency parser and all the variant of our parser in three settings:
		considering all constituents, considering constituents with a block degree less or equal to two and exactly two.
		For the last case, we also report the number of constituents.
		We do not remove punctuation.
		The analysis is done with the concatenation of train, development and test sets.
	}
	\label{table:cover}
\end{table*}
\begin{table*}[p]
	\centering
	\scriptsize
	\begin{tabular}{@{}lcccccccc@{}}
		\toprule
		& \multicolumn{2}{c}{\bf \textsc{\negra{}}}
		& \phantom{abc}
		& \multicolumn{2}{c}{\bf \textsc{\tiger{}}}
		& \phantom{abc}
		& \multicolumn{2}{c}{\bf {\dptb{}}}
		\\
		\cmidrule{2-3} \cmidrule{5-6} \cmidrule{8-9}
		& F1 & Disc. F1 &
		& F1 & Disc. F1 &
		& F1 & Disc. F1
		\\
		\midrule
\multicolumn{9}{l}{\bf Fully supervised} \\
\midrule
\newcite{gonzalez2015reduction}
& 77.0 & &
& & &
& 77.3 &
\\
\newcite{versley2016discontinuity}
&  & &
& 79.5 & &
&  &
\\
\newcite{gebhardt2018generic}
&  & &
& 75.1 & &
&  &
\\
\newcite{corro2017gmsa}
&  & &
& & &
& 89.2 &
\\
\newcite{coavoux2017incremental}
&  & &
& 79.3 & &
&  &
\\
\newcite{coavoux2019unlex}
& 83.2  & 54.6 &
& 82.7 & 55.9 &
& 91.0 & {\bf 71.3}
\\
\newcite{coavoux2019stackfree}
& 83.2 & {\bf 56.3} &
& 82.5 & 55.9 &
& 90.9 & 67.3
\\

\newcite{fernandez2020pointer}
& 83.7 & 54.7 &
& 84.6 &  {\bf 57.9} &
&  &
\\
\rowcolor{gray!20}
This work, $\mathcal O(n^3)$
& {\bf 86.2} & 54.1 & 
& {\bf 85.5} & 53.8 & 
& {\bf 92.7} & 64.2 

\\
\rowcolor{gray!20}
This work, $\mathcal O(n^5)$ and $\mathcal O(n^6)$, well-nested
& 84.9 & 46.1 & 
& 84.8 & 50.4 & 
& 92.6 & 62.6 
\\
\rowcolor{gray!20}
This work, $\mathcal O(n^5)$ and $\mathcal O(n^6)$
& 84.9 & 46.2 & 
& 84.9 & 51.0 & 
& 92.6 & 62.9 
\\
\\
		\multicolumn{9}{l}{\bf + gold part-of-speech tags} \\
\newcite{maier2015shiftreduce}
& 77.0 & 19.8 &
& 74.7 & 18.8 &
&  &
\\
\newcite{coavoux2017incremental}
& 82.2 & 50.0 &
& 81.6 & 49.2 &
&  &
\\
\newcite{corro2017gmsa}
&  & &
&81.6 & &
& 90.1 &
\\
\\
		\midrule
\multicolumn{9}{l}{\bf Semi-supervised: pre-trained word embeddings} \\
\midrule
\newcite{stanojevic2017neural}
&  & &
& 77.0 & &
&  &
\\
\newcite{fernandez2020pointer}, with pred tags
& 85.4 &{\bf 58.8} &
& 85.3 & 59.1 &
&  &
\\
\newcite{fernandez2020pointer}, without pred tags
& 85.7 & 58.6 &
& {\bf 85.7} & {\bf 60.4} &
&  &
\\
\rowcolor{gray!20}
This work, $\mathcal O(n^3)$
& {\bf 86.3} & 56.1 & 
& 85.2 & 51.2 & 
& {\bf 92.9} & {\bf 64.9} 
\\
\rowcolor{gray!20}
This work, $\mathcal O(n^5)$ and $\mathcal O(n^6)$, well-nested
& 85.6 & 52.9 & 
& 84.9 & 50.4 & 
& 92.6 & 59.4 
\\
\rowcolor{gray!20}
This work, $\mathcal O(n^5)$ and $\mathcal O(n^6)$
& 85.6 & 53.0 & 
& 84.9 & 51.0 & 
& 92.6 &  59.7 
\\
\\
		\multicolumn{9}{l}{\bf + gold POS tags} \\
\newcite{stanojevic2017neural}
& 82.9 & &
& 81.6 & &
&  &
\\
\newcite{fernandez2020pointer}
& 86.1 & 59.9 &
& 86.3 & 60.7 &
&  &
\\
\midrule
		\multicolumn{9}{l}{\bf Semi-supervised: \bert{}} \\
\midrule
\rowcolor{gray!20}
This work, $\mathcal O(n^3)$
& {\bf 91.6} & {\bf 66.1} & 
& {\bf 90.0} & {\bf 62.1} & 
& {\bf 94.8} & {\bf 68.9} 
\\
\rowcolor{gray!20}
This work, $\mathcal O(n^5)$ and $\mathcal O(n^6)$, well-nested
& 90.5 & 58.8 & 
& 89.3 & 57.8 & 
& 94.5 & 64.5 
\\
\rowcolor{gray!20}
This work, $\mathcal O(n^5)$ and $\mathcal O(n^6)$
& 90.6 & 59.6 & 
& 89.3 & 58.7 & 
& 94.5 & 64.7 
\\
		\bottomrule
	\end{tabular}
    \caption{%
    	Discontinuous constituency parsing results on the three test sets.
    	The $\mathcal O(n^5)$ and $\mathcal O(n^6)$ variants produced exactly the same results in all settings.
    }
    \label{table:results}
\end{table*}

\begin{table*}[p]
	\centering
	\scriptsize
	\begin{tabular}{@{}lccccccccccc@{}}
		\toprule
		& \multicolumn{3}{c}{\bf \textsc{\negra{}}}
		& 
		& \multicolumn{3}{c}{\bf \textsc{\tiger{}}}
		& 
		& \multicolumn{3}{c}{\bf {\dptb{}}}
		\\
		\cmidrule{2-4} \cmidrule{6-8} \cmidrule{10-12}
		& D. recall & D. prec. & D. F1 &
		& D. recall & D. prec. & D. F1 & 
		& D. recall & D. prec. & D. F1
		\\
		\midrule
		$\mathcal O(n^3)$
		& 42.0 & {\bf 76.0} & {\bf 54.1} & 
		& 40.6 & {\bf 79.8} & {\bf 53.8} & 
		& 49.7 & {\bf 90.8} & {\bf 64.2}
		\\
		$\mathcal O(n^5)$ and $\mathcal O(n^6)$, wn
		& 47.0 & 45.2 & 46.1 & 
		& 46.9 &54.3 & 50.4 & 
		& 63.8 &61.4 & 62.6
		\\
		$\mathcal O(n^5)$and $\mathcal O(n^6)$
		& {\bf 47.3} & 45.2 & 46.2 & 
		& {\bf 47.8} &54.8 & 51.0 & 
		& {\bf 64.0} & 61.8  & 62.9
		\\
		\bottomrule
	\end{tabular}
    \caption{%
    	Detailed discontinuous constituency parsing results for the fully supervised model.
    }
    \label{table:detailed_results}
\end{table*}

\subsection{Neural parameterization}

We use a neural architecture based on bidirectional LSTMs detailed in Appendix~\ref{app:neural}.

{\bf Constituent scores}
Even with the block degree two restriction, there is a larger number (quartic!) of constituent scores to compute.
In early experiments, we observed that weighting such a number of constituents without further decomposition blow up the neural network memory usage and was prohibitively slow.
Therefore, we introduce a score decomposition that results in a quadratic number of scores to compute and that can be efficiently parallelized on GPU using batched matrix operations.%
\footnote{Technical details about the implementation are out of the scope of this paper. However, they are described in our implementation that will be publicly available.}

We decompose the score of a constituent $[A, i, k, l, j]$ as the sum of a score associated with its \underline{o}uter boundaries (i.e. indices $i$ and $j$) and one with its \underline{g}ap boundaries (i.e. indices $k$ and $l$).
The score of constituent  is defined as:\footnote{The +1 in tensor indices result of the fact that we use interstice indices for constituents but that the neural network layers focus on word indices.}
\begin{align*}
\etW_{A, i, k, l, j} = \begin{cases}
\etS^{\text{(c. label)}}_{A, i+1, j} + \emS^{\text{(c. span)}}_{i+1, j}  &\text{ if } i = k = -, \\[10pt]
\etS^{\text{(o. label)}}_{A, i+1, j} + \emS^{\text{(o. span)}}_{i+1, j} + \etS^{\text{(g. label)}}_{A, k+1, l} + \emS^{\text{(g. span)}}_{k + 1}  &\text{ otherwise}.
\end{cases}
\end{align*}
where tensors $\tS^{\text{(c. label)}}$, $\tS^{\text{(o. label)}}$, $\tS^{\text{(g. label)}}~\in~\mathbb R^{|\mathcal{V}| \times n \times n}$
and matrices $\mS^{\text{(c. span)}}$, $\mS^{\text{(o. span)}}$, $\mS^{\text{(g. span)}}~\in~\mathbb R^{n \times n}$ are computed using the deep biaffine attention mechanism \cite{dozat2016biaffine}.
The tensor $\tW$ is never explicitly built: during the dynamic program execution we lazily compute requested constituent scores.

{\bf Training loss}
Span-based continuous constituency parsers are usually trained using a decomposable margin-based objective \cite{stern2017spanbased,kitaev2018parser,kitaev2019bertparser}.
This approach requires to repeatedly perform loss-augmented inference during training \cite{taskar2005lossaugmented}, which can be prohibitively slow even when tractable.
A current trend in dependency parsing is to ignore the tree structure and rely on negative log likelihood for head selection for each modifier word independently \cite{dozat2016biaffine,zhang2017headselection}.
We rely on a similar approach and use as training objective the negative log-likelihood loss independently for each span (continuous, outer and gap), adding a null label with a fixed 0 weight as label for spans that do not appear in the gold annotation.

\begin{table*}
	\centering
	\small
	\begin{tabular}{@{}crcccccc@{}}
		\toprule
		& nn &  $\mathcal O(n^3)$ &  $\mathcal O(n^4)$ &  $\mathcal O(n^5)$, wn &  $\mathcal O(n^5)$ &  $\mathcal O(n^6)$, wn &  $\mathcal O(n^6)$ \\
		\hline
		\negra{} & 1.74 & 0.35 & 1.10 & 3.73 &  4.48 & 8.82 & 22.58  \\
		\tiger{} & 7.73 & 2.81 & 12.96 & 98.44 & 133.22 & 507.84 & 1841.95  \\
		\dptb{} & 4.67 & 2.13 & 6.70 & 19.35 & 22.71 & 43.00 & 105.98  \\
		\bottomrule
	\end{tabular}
	\caption{%
		Total time in seconds to parse the full test sets:
		the nn column corresponds to the time taken by the forward pass of the neural network (maximum 5000 words per batch on a NVIDIA Tesla V100 SXM2 32 Go),
		each supplementary column is the time take by each variant of the chart-based algorithm (without any parallelization, one sentence at a time).
	}
	\label{table:timing}
\end{table*}

\subsection{Evaluation}

We evaluate our parser on the test sets of the three treebanks.
We report F-measure and discontinuous F-measure as computed using the disco-dop tool\footnote{\url{https://github.com/andreasvc/disco-dop}} with the standard parameters in Table~\ref{table:results}.

First, we observed that the $\mathcal O(n^5)$ and $\mathcal O(n^6)$ variants of our parsers produced exactly the same results in all settings.
This may be expected as their cover of the original treebanks where almost similar.
Second, surprisingly, the $\mathcal O(n^3)$ parser produced better results in term of F-measure than other variants in all cases.
We report labeled discontinuous constituent recall and precision measures for the fully supervised model in Table~\ref{table:detailed_results}.
We observe that while the $\mathcal O(n^5)$ and $\mathcal O(n^6)$ have an better recall than the $\mathcal O(n^3)$ parser, their precision is drastically lower.
This highlights a benefit of restricting the search space: the parser can retrieve less erroneous constituents leading to an improved overall precision.

Finally, in almost all cases, we achieve a novel state-of-the-art for the task in term of labeled F-measure.
However, we are slightly lower when evaluating discontinuous constituent only.
We suspect that this is due to the fact that our best parser is the one with the smallest search space.

\subsection{Runtime}

The runtime on the test sets of our approach is reported on Table~\ref{table:timing}.
In all cases, the runtime is reasonably fast and we do not need to remove long sentences.
Interestingly, most of the time is spent for computing scores with the neural network with the cubic time parser, even if we use batches to benefit from the GPU architecture while our chart-based algorithm is not paralellized on CPU.
\section{Conclusion}

We proposed a novel family of algorithms for discontinuous constituency parsing achieving state-of-the art results.
Importantly, we showed that a specific set of discontinuous constituency trees can be parsed in cubic time while covering most of the linguistict structures observed in treebanks.
Despite being based on chart-based algorithms, our approach is fast as test time and we can parse all sentences without pruning or filtering long sentences.
Future research could explore neural architecture and training losses tailored to this approach for discontinuous constituency parsing.

\section*{Acknowledgments}
We thank Djam\'e Seddah, Maximin Coavoux and Vlad Niculae for their comments and suggestions.
We thank Serhii Havrylov for the help with the code.
We thank Ivan Titov for comments on the early draft of this paper.
We thank  Laura  Kallmeyer and Kilian Evang for providing us with the script for  the  discontinuous  PTB.
We thank Maximin Coavoux for  the  help  with  pre-processing the data.
The project was supported by the Dutch National Science Foundation (NWO VIDI 639.022.518) and European Research Council (ERC Starting Grant BroadSem 678254).
This work benefited from the Jean-Zay cluster.

\bibliographystyle{coling}
\bibliography{tacl2018}

\begin{thebibliography}{}

\bibitem[\protect\citename{Bangalore and Joshi}1999]{bangalore1999supertagging}
Srinivas Bangalore and Aravind~K. Joshi.
\newblock 1999.
\newblock {S}upertagging: An approach to almost parsing.
\newblock {\em Computational Linguistics}, 25(2):237--265.

\bibitem[\protect\citename{Boullier}1998]{boullier1998proposal}
Pierre Boullier.
\newblock 1998.
\newblock Proposal for a natural language processing syntactic backbone.

\bibitem[\protect\citename{Boullier}2004]{boullier2004rcg}
Pierre Boullier.
\newblock 2004.
\newblock Range concatenation grammars.
\newblock In {\em New developments in parsing technology}, pages 269--289.
  Springer.

\bibitem[\protect\citename{Brants \bgroup et al.\egroup }2002]{brants2002tiger}
Sabine Brants, Stefanie Dipper, Silvia Hansen, Wolfgang Lezius, and George
  Smith.
\newblock 2002.
\newblock The tiger treebank.
\newblock In {\em Proceedings of the workshop on treebanks and linguistic
  theories}, volume 168.

\bibitem[\protect\citename{Bunt \bgroup et al.\egroup
  }1987]{bunt1987discontinuous}
Harry Bunt, Jan Thesingh, and Ko~van~der Sloot.
\newblock 1987.
\newblock Discontinuous constituents in trees, rules, and parsing.
\newblock In {\em Third Conference of the {E}uropean Chapter of the Association
  for Computational Linguistics}, Copenhagen, Denmark, April. Association for
  Computational Linguistics.

\bibitem[\protect\citename{Chen-Main and Joshi}2010]{chen2010illnestedness}
Joan Chen-Main and Aravind~K. Joshi.
\newblock 2010.
\newblock Unavoidable ill-nestedness in natural language and the adequacy of
  tree local-{MCTAG} induced dependency structures.
\newblock In {\em Proceedings of the 10th International Workshop on Tree
  Adjoining Grammar and Related Frameworks ({TAG}+10)}, pages 53--60, Yale
  University, June. Linguistic Department, Yale University.

\bibitem[\protect\citename{Coavoux and Cohen}2019]{coavoux2019stackfree}
Maximin Coavoux and Shay~B. Cohen.
\newblock 2019.
\newblock Discontinuous constituency parsing with a stack-free transition
  system and a dynamic oracle.
\newblock In {\em Proceedings of the 2019 Conference of the North {A}merican
  Chapter of the Association for Computational Linguistics: Human Language
  Technologies, Volume 1 (Long and Short Papers)}, pages 204--217, Minneapolis,
  Minnesota, June. Association for Computational Linguistics.

\bibitem[\protect\citename{Coavoux and Crabb{\'e}}2017]{coavoux2017incremental}
Maximin Coavoux and Beno{\^\i}t Crabb{\'e}.
\newblock 2017.
\newblock Incremental discontinuous phrase structure parsing with the {GAP}
  transition.
\newblock In {\em Proceedings of the 15th Conference of the {E}uropean Chapter
  of the Association for Computational Linguistics: Volume 1, Long Papers},
  pages 1259--1270, Valencia, Spain, April. Association for Computational
  Linguistics.

\bibitem[\protect\citename{Coavoux \bgroup et al.\egroup
  }2019]{coavoux2019unlex}
Maximin Coavoux, Benoît Crabbé, and Shay~B. Cohen.
\newblock 2019.
\newblock Unlexicalized transition-based discontinuous constituency parsing.
\newblock {\em Transactions of the Association for Computational Linguistics},
  7:73--89.

\bibitem[\protect\citename{Cocke}1969]{Cocke1969cyk}
John Cocke.
\newblock 1969.
\newblock {\em Programming Languages and Their Compilers: Preliminary Notes}.
\newblock New York University, New York, NY, USA.

\bibitem[\protect\citename{Corro \bgroup et al.\egroup }2016]{corro2016bdwn}
Caio Corro, Joseph Le~Roux, Mathieu Lacroix, Antoine Rozenknop, and Roberto
  Wolfler~Calvo.
\newblock 2016.
\newblock Dependency parsing with bounded block degree and well-nestedness via
  {L}agrangian relaxation and branch-and-bound.
\newblock In {\em Proceedings of the 54th Annual Meeting of the Association for
  Computational Linguistics (Volume 1: Long Papers)}, pages 355--366, Berlin,
  Germany, August. Association for Computational Linguistics.

\bibitem[\protect\citename{Corro \bgroup et al.\egroup }2017]{corro2017gmsa}
Caio Corro, Joseph Le~Roux, and Mathieu Lacroix.
\newblock 2017.
\newblock Efficient discontinuous phrase-structure parsing via the generalized
  maximum spanning arborescence.
\newblock In {\em Proceedings of the 2017 Conference on Empirical Methods in
  Natural Language Processing}, pages 1644--1654, Copenhagen, Denmark,
  September. Association for Computational Linguistics.

\bibitem[\protect\citename{Cross and Huang}2016]{cross2016span}
James Cross and Liang Huang.
\newblock 2016.
\newblock Span-based constituency parsing with a structure-label system and
  provably optimal dynamic oracles.
\newblock In {\em Proceedings of the 2016 Conference on Empirical Methods in
  Natural Language Processing}, pages 1--11, Austin, Texas, November.
  Association for Computational Linguistics.

\bibitem[\protect\citename{Dozat and Manning}2016]{dozat2016biaffine}
Timothy Dozat and Christopher~D Manning.
\newblock 2016.
\newblock Deep biaffine attention for neural dependency parsing.
\newblock {\em arXiv preprint arXiv:1611.01734}.

\bibitem[\protect\citename{Dubey and Keller}2003]{dubey2003negra}
Amit Dubey and Frank Keller.
\newblock 2003.
\newblock Probabilistic parsing for german using sister-head dependencies.
\newblock In {\em Proceedings of the 41st Annual Meeting on Association for
  Computational Linguistics-Volume 1}, pages 96--103. Association for
  Computational Linguistics.

\bibitem[\protect\citename{Eisner and Satta}1999]{eisner1999bilexical}
Jason Eisner and Giorgio Satta.
\newblock 1999.
\newblock Efficient parsing for bilexical context-free grammars and head
  automaton grammars.
\newblock In {\em Proceedings of the 37th Annual Meeting of the Association for
  Computational Linguistics}, pages 457--464, College Park, Maryland, USA,
  June. Association for Computational Linguistics.

\bibitem[\protect\citename{Eisner and Satta}2000]{eisner2000fasterltag}
Jason Eisner and Giorgio Satta.
\newblock 2000.
\newblock A faster parsing algorithm for lexicalized tree-adjoining grammars.
\newblock In {\em Proceedings of the Fifth International Workshop on Tree
  Adjoining Grammar and Related Frameworks ({TAG}+5)}, pages 79--84,
  Universit{\'e} Paris 7, May.

\bibitem[\protect\citename{Eisner}1997]{eisner1997bilexical}
Jason Eisner.
\newblock 1997.
\newblock Bilexical grammars and a cubic-time probabilistic parser.
\newblock In {\em Advances in probabilistic and other parsing technologies},
  pages 54--–65.

\bibitem[\protect\citename{Evang and Kallmeyer}2011]{evang2011plcfrs}
Kilian Evang and Laura Kallmeyer.
\newblock 2011.
\newblock {PLCFRS} parsing of {E}nglish discontinuous constituents.
\newblock In {\em Proceedings of the 12th International Conference on Parsing
  Technologies}, pages 104--116, Dublin, Ireland, October. Association for
  Computational Linguistics.

\bibitem[\protect\citename{Fern{\'a}ndez-Gonz{\'a}lez and
  Martins}2015]{gonzalez2015reduction}
Daniel Fern{\'a}ndez-Gonz{\'a}lez and Andr{\'e} F.~T. Martins.
\newblock 2015.
\newblock Parsing as reduction.
\newblock In {\em Proceedings of the 53rd Annual Meeting of the Association for
  Computational Linguistics and the 7th International Joint Conference on
  Natural Language Processing (Volume 1: Long Papers)}, pages 1523--1533,
  Beijing, China, July. Association for Computational Linguistics.

\bibitem[\protect\citename{Fernández-González and
  Gómez-Rodríguez}2020]{fernandez2020pointer}
Daniel Fernández-González and Carlos Gómez-Rodríguez.
\newblock 2020.
\newblock Discontinuous constituent parsing with pointer networks.

\bibitem[\protect\citename{Gebhardt}2018]{gebhardt2018generic}
Kilian Gebhardt.
\newblock 2018.
\newblock Generic refinement of expressive grammar formalisms with an
  application to discontinuous constituent parsing.
\newblock In {\em Proceedings of the 27th International Conference on
  Computational Linguistics}, pages 3049--3063, Santa Fe, New Mexico, USA,
  August. Association for Computational Linguistics.

\bibitem[\protect\citename{G{\'o}mez-Rodr{\'\i}guez \bgroup et al.\egroup
  }2009]{gomez2009nonproj}
Carlos G{\'o}mez-Rodr{\'\i}guez, David Weir, and John Carroll.
\newblock 2009.
\newblock Parsing mildly non-projective dependency structures.
\newblock In {\em Proceedings of the 12th Conference of the {E}uropean Chapter
  of the {ACL} ({EACL} 2009)}, pages 291--299, Athens, Greece, March.
  Association for Computational Linguistics.

\bibitem[\protect\citename{G{\'o}mez-Rodr{\'\i}guez \bgroup et al.\egroup
  }2011]{gomez2011nonproj}
Carlos G{\'o}mez-Rodr{\'\i}guez, John Carroll, and David Weir.
\newblock 2011.
\newblock Dependency parsing schemata and mildly non-projective dependency
  parsing.
\newblock {\em Computational Linguistics}, 37(3):541--586.

\bibitem[\protect\citename{Goodman}1999]{goodman1999semiring}
Joshua Goodman.
\newblock 1999.
\newblock Semiring parsing.
\newblock {\em Computational Linguistics}, 25(4):573--606.

\bibitem[\protect\citename{Hall \bgroup et al.\egroup }2014]{hall2014spanbased}
David Hall, Greg Durrett, and Dan Klein.
\newblock 2014.
\newblock Less grammar, more features.
\newblock In {\em Proceedings of the 52nd Annual Meeting of the Association for
  Computational Linguistics (Volume 1: Long Papers)}, pages 228--237,
  Baltimore, Maryland, June. Association for Computational Linguistics.

\bibitem[\protect\citename{Kallmeyer and Maier}2010]{kallmeyer2010lcfrs}
Laura Kallmeyer and Wolfgang Maier.
\newblock 2010.
\newblock Data-driven parsing with probabilistic linear context-free rewriting
  systems.
\newblock In {\em Proceedings of the 23rd International Conference on
  Computational Linguistics (Coling 2010)}, pages 537--545, Beijing, China,
  August. Coling 2010 Organizing Committee.

\bibitem[\protect\citename{Kasami}1966]{kasami1966cyk}
Tadao Kasami.
\newblock 1966.
\newblock An efficient recognition and syntax-analysis algorithm for
  context-free languages.
\newblock {\em Coordinated Science Laboratory Report no. R-257}.

\bibitem[\protect\citename{Kiperwasser and
  Goldberg}2016]{kiperwasser2016parser}
Eliyahu Kiperwasser and Yoav Goldberg.
\newblock 2016.
\newblock Simple and accurate dependency parsing using bidirectional {LSTM}
  feature representations.
\newblock {\em Transactions of the Association for Computational Linguistics},
  4:313--327.

\bibitem[\protect\citename{Kitaev and Klein}2018]{kitaev2018parser}
Nikita Kitaev and Dan Klein.
\newblock 2018.
\newblock Constituency parsing with a self-attentive encoder.
\newblock In {\em Proceedings of the 56th Annual Meeting of the Association for
  Computational Linguistics (Volume 1: Long Papers)}, pages 2676--2686,
  Melbourne, Australia, July. Association for Computational Linguistics.

\bibitem[\protect\citename{Kitaev \bgroup et al.\egroup
  }2019]{kitaev2019bertparser}
Nikita Kitaev, Steven Cao, and Dan Klein.
\newblock 2019.
\newblock Multilingual constituency parsing with self-attention and
  pre-training.
\newblock In {\em Proceedings of the 57th Annual Meeting of the Association for
  Computational Linguistics}, pages 3499--3505, Florence, Italy, July.
  Association for Computational Linguistics.

\bibitem[\protect\citename{Kuhlmann and Nivre}2006]{kuhlmann2006mildly}
Marco Kuhlmann and Joakim Nivre.
\newblock 2006.
\newblock Mildly non-projective dependency structures.
\newblock In {\em Proceedings of the {COLING}/{ACL} 2006 Main Conference Poster
  Sessions}, pages 507--514, Sydney, Australia, July. Association for
  Computational Linguistics.

\bibitem[\protect\citename{Maier and Kallmeyer}2010]{maier2010nonproj}
Wolfgang Maier and Laura Kallmeyer.
\newblock 2010.
\newblock Discontinuity and non-projectivity: Using mildly context-sensitive
  formalisms for data-driven parsing.
\newblock In {\em Proceedings of the 10th International Workshop on Tree
  Adjoining Grammar and Related Frameworks ({TAG}+10)}, pages 119--126, Yale
  University, June. Linguistic Department, Yale University.

\bibitem[\protect\citename{Maier and Lichte}2016]{maier2016discontinuous}
Wolfgang Maier and Timm Lichte.
\newblock 2016.
\newblock Discontinuous parsing with continuous trees.
\newblock In {\em Proceedings of the Workshop on Discontinuous Structures in
  Natural Language Processing}, pages 47--57, San Diego, California, June.
  Association for Computational Linguistics.

\bibitem[\protect\citename{Maier \bgroup et al.\egroup }2012]{maier2012plcfrs}
Wolfgang Maier, Miriam Kaeshammer, and Laura Kallmeyer.
\newblock 2012.
\newblock {PLCFRS} parsing revisited: Restricting the fan-out to two.
\newblock In {\em Proceedings of the 11th International Workshop on Tree
  Adjoining Grammars and Related Formalisms ({TAG}+11)}, pages 126--134, Paris,
  France, September.

\bibitem[\protect\citename{Maier}2015]{maier2015shiftreduce}
Wolfgang Maier.
\newblock 2015.
\newblock Discontinuous incremental shift-reduce parsing.
\newblock In {\em Proceedings of the 53rd Annual Meeting of the Association for
  Computational Linguistics and the 7th International Joint Conference on
  Natural Language Processing (Volume 1: Long Papers)}, pages 1202--1212,
  Beijing, China, July. Association for Computational Linguistics.

\bibitem[\protect\citename{Marcus \bgroup et al.\egroup }1993]{marcus1993ptb}
Mitchell Marcus, Beatrice Santorini, and Mary~Ann Marcinkiewicz.
\newblock 1993.
\newblock Building a large annotated corpus of english: The penn treebank.

\bibitem[\protect\citename{McCawley}1982]{mccawley1982discontinuous}
James~D. McCawley.
\newblock 1982.
\newblock Parentheticals and discontinuous constituent structure.
\newblock {\em Linguistic Inquiry}, 13(1):91--106.

\bibitem[\protect\citename{McDonald \bgroup et al.\egroup
  }2005]{mcdonald2005mst}
Ryan McDonald, Fernando Pereira, Kiril Ribarov, and Jan Haji{\v{c}}.
\newblock 2005.
\newblock Non-projective dependency parsing using spanning tree algorithms.
\newblock In {\em Proceedings of Human Language Technology Conference and
  Conference on Empirical Methods in Natural Language Processing}, pages
  523--530, Vancouver, British Columbia, Canada, October. Association for
  Computational Linguistics.

\bibitem[\protect\citename{M{\"u}ller}2004]{muller2004continuous}
Stefan M{\"u}ller.
\newblock 2004.
\newblock Continuous or discontinuous constituents? a comparison between
  syntactic analyses for constituent order and their processing systems.
\newblock {\em Research on Language and Computation}, 2(2):209--257.

\bibitem[\protect\citename{Myung \bgroup et al.\egroup }1995]{myung1995gmsa}
Young-Soo Myung, Chang-Ho Lee, and Dong-Wan Tcha.
\newblock 1995.
\newblock On the generalized minimum spanning tree problem.
\newblock {\em Networks}, 26(4):231--241.

\bibitem[\protect\citename{Nivre}2009]{nivre2009swap}
Joakim Nivre.
\newblock 2009.
\newblock Non-projective dependency parsing in expected linear time.
\newblock In {\em Proceedings of the Joint Conference of the 47th Annual
  Meeting of the {ACL} and the 4th International Joint Conference on Natural
  Language Processing of the {AFNLP}}, pages 351--359, Suntec, Singapore,
  August. Association for Computational Linguistics.

\bibitem[\protect\citename{Pereira and Warren}1983]{pereira1983deduction}
Fernando C.~N. Pereira and David H.~D. Warren.
\newblock 1983.
\newblock Parsing as deduction.
\newblock In {\em 21st Annual Meeting of the Association for Computational
  Linguistics}, pages 137--144, Cambridge, Massachusetts, USA, June.
  Association for Computational Linguistics.

\bibitem[\protect\citename{Pitler \bgroup et al.\egroup
  }2012]{pitler2012gapminding}
Emily Pitler, Sampath Kannan, and Mitchell Marcus.
\newblock 2012.
\newblock Dynamic programming for higher order parsing of gap-minding trees.
\newblock In {\em Proceedings of the 2012 Joint Conference on Empirical Methods
  in Natural Language Processing and Computational Natural Language Learning},
  pages 478--488, Jeju Island, Korea, July. Association for Computational
  Linguistics.

\bibitem[\protect\citename{Rambow}2010]{rambow2010opinion}
Owen Rambow.
\newblock 2010.
\newblock The simple truth about dependency and phrase structure
  representations: An opinion piece.
\newblock In {\em Human Language Technologies: The 2010 Annual Conference of
  the North {A}merican Chapter of the Association for Computational
  Linguistics}, pages 337--340, Los Angeles, California, June. Association for
  Computational Linguistics.

\bibitem[\protect\citename{Satta and Kuhlmann}2013]{satta2013headsplit}
Giorgio Satta and Marco Kuhlmann.
\newblock 2013.
\newblock Efficient parsing for head-split dependency trees.
\newblock {\em Transactions of the Association for Computational Linguistics},
  1:267--278.

\bibitem[\protect\citename{Saxe \bgroup et al.\egroup
  }2013]{saxe2013othogonal_init}
Andrew~M. Saxe, James~L. McClelland, and Surya Ganguli.
\newblock 2013.
\newblock Exact solutions to the nonlinear dynamics of learning in deep linear
  neural networks.

\bibitem[\protect\citename{Seddah \bgroup et al.\egroup }2014]{seddah2014spmrl}
Djam{\'e} Seddah, Sandra K{\"u}bler, and Reut Tsarfaty.
\newblock 2014.
\newblock Introducing the spmrl 2014 shared task on parsing
  morphologically-rich languages.
\newblock In {\em Proceedings of the First Joint Workshop on Statistical
  Parsing of Morphologically Rich Languages and Syntactic Analysis of
  Non-Canonical Languages}, pages 103--109.

\bibitem[\protect\citename{Seki \bgroup et al.\egroup }1991]{seki1991mcfg}
Hiroyuki Seki, Takashi Matsumura, Mamoru Fujii, and Tadao Kasami.
\newblock 1991.
\newblock On multiple context-free grammars.
\newblock {\em Theoretical Computer Science}, 88(2):191--229.

\bibitem[\protect\citename{Skut \bgroup et al.\egroup }1997]{skut1997negra}
Wojciech Skut, Brigitte Krenn, Thorsten Brants, and Hans Uszkoreit.
\newblock 1997.
\newblock An annotation scheme for free word order languages.
\newblock {\em arXiv preprint cmp-lg/9702004}.

\bibitem[\protect\citename{Stanojevi{\'c} and
  Alhama}2017]{stanojevic2017neural}
Milo{\v{s}} Stanojevi{\'c} and Raquel~G. Alhama.
\newblock 2017.
\newblock Neural discontinuous constituency parsing.
\newblock In {\em Proceedings of the 2017 Conference on Empirical Methods in
  Natural Language Processing}, pages 1666--1676, Copenhagen, Denmark,
  September. Association for Computational Linguistics.

\bibitem[\protect\citename{Stanojevi{\'c} and
  G.~Alhama}2017]{stanojevic2017neuralshift}
Milo{\v{s}} Stanojevi{\'c} and Raquel G.~Alhama.
\newblock 2017.
\newblock Neural discontinuous constituency parsing.
\newblock In {\em Proceedings of the 2017 Conference on Empirical Methods in
  Natural Language Processing}, pages 1666--1676, Copenhagen, Denmark,
  September. Association for Computational Linguistics.

\bibitem[\protect\citename{Stern \bgroup et al.\egroup
  }2017]{stern2017spanbased}
Mitchell Stern, Jacob Andreas, and Dan Klein.
\newblock 2017.
\newblock A minimal span-based neural constituency parser.
\newblock In {\em Proceedings of the 55th Annual Meeting of the Association for
  Computational Linguistics (Volume 1: Long Papers)}, pages 818--827,
  Vancouver, Canada, July. Association for Computational Linguistics.

\bibitem[\protect\citename{Taskar \bgroup et al.\egroup
  }2005]{taskar2005lossaugmented}
Ben Taskar, Vassil Chatalbashev, Daphne Koller, and Carlos Guestrin.
\newblock 2005.
\newblock Learning structured prediction models: A large margin approach.
\newblock In {\em Proceedings of the 22nd international conference on Machine
  learning}, pages 896--903. ACM.

\bibitem[\protect\citename{Versley}2014]{versley2014swapconst}
Yannick Versley.
\newblock 2014.
\newblock Experiments with easy-first nonprojective constituent parsing.
\newblock In {\em Proceedings of the First Joint Workshop on Statistical
  Parsing of Morphologically Rich Languages and Syntactic Analysis of
  Non-Canonical Languages}, pages 39--53, Dublin, Ireland, August. Dublin City
  University.

\bibitem[\protect\citename{Versley}2016]{versley2016discontinuity}
Yannick Versley.
\newblock 2016.
\newblock Discontinuity (re){\mbox{$^2$}}-visited: A minimalist approach to
  pseudoprojective constituent parsing.
\newblock In {\em Proceedings of the Workshop on Discontinuous Structures in
  Natural Language Processing}, pages 58--69, San Diego, California, June.
  Association for Computational Linguistics.

\bibitem[\protect\citename{Vijay-Shanker \bgroup et al.\egroup
  }1987]{vijay-shanker1987lcfrs}
K.~Vijay-Shanker, David~J. Weir, and Aravind~K. Joshi.
\newblock 1987.
\newblock Characterizing structural descriptions produced by various
  grammatical formalisms.
\newblock In {\em 25th Annual Meeting of the Association for Computational
  Linguistics}, pages 104--111, Stanford, California, USA, July. Association
  for Computational Linguistics.

\bibitem[\protect\citename{Younger}1967]{younger1967cyk}
Daniel~H Younger.
\newblock 1967.
\newblock Recognition and parsing of context-free languages in time n3.
\newblock {\em Information and control}, 10(2):189--208.

\bibitem[\protect\citename{Zhang \bgroup et al.\egroup
  }2017]{zhang2017headselection}
Xingxing Zhang, Jianpeng Cheng, and Mirella Lapata.
\newblock 2017.
\newblock Dependency parsing as head selection.
\newblock In {\em Proceedings of the 15th Conference of the {E}uropean Chapter
  of the Association for Computational Linguistics: Volume 1, Long Papers},
  pages 665--676, Valencia, Spain, April. Association for Computational
  Linguistics.

\end{thebibliography}

\section{Neural parameterization}
\label{app:neural}
In this appendix, we describe the different components of our neural network.
If unspecified, parameters are initialized with Pytorch default initialization.

\subsection{Word-level features}

We use three kind of word-level features:
word embeddings, character embeddings and, for a few experiments, part-of-speech embeddings.
All embeddings are concatenated to form word-level embeddings.

Word embeddings can either be pre-trained or trained end-to-end.
In the case of pre-trained word embeddings, we fix them and sum them with end-to-end learned word embeddings initialized at 0.

Character embeddings are fed to a BiLSTM.
The hidden states of the two endpoints are then concatenated together.
Words are truncated to 20 characters for this feature.

\subsection{Sentence-level features}

We follow \cite{kiperwasser2016parser}  by using two stacked BiLSTM, i.e. the input of the second BiLSTM is the concatenation of the forward and backward hidden states of the first one.
All LSTMs have a single layer.
Projection matrices are initialized with the orthogonal approach proposed by \cite{saxe2013othogonal_init} and bias vectors are initialized to 0.

For models using Bert, we learn a convex combination of the last 4 layers, in a similar spirit to ELMO.
When word are tokenized in subwords by the Bert tokenizer, we use the embedding of the first sub-token.

\subsection{Output weights}

We have two different output layers.
First, we predict part-of-speech tags with a linear projection on top of the hidden states of the first BiLSTM.
During training, we use an auxiliary negative log-likelihood loss.
Second, after the second BiLSTM we add the biaffine layers to compute span scores.

\subsection{Hyperparameters}

We report the dimensions of the building blocks of the network in Table~\ref{table:hyperparams}.

We optimize the parameters with the Adam variant of stochastic gradient descent descent with minibatches containing at most 5000 words for 200 epochs.
We apply dropout with ratio $0.3$ before the input of the character BiLSTM, before the first stack of sentence-level BiLSTM and after the second one by following the methodology of \newcite{dozat2016biaffine}.

\begin{table*}
	\begin{tabular}{l|c}
    Name & Dimension \\
    \hline
    Word embeddings & 300 \\
    Character embeddings & 64 \\
    Character BiLSTM & 100 \\
    Character BiLSTMs layer & 1 \\
    Sentence BiLSTMs & 800 \\
    Sentence BiLSTMs layer & 1 \\
    Sentence BiLSTMs stack & 2 \\
    Span projection & 500 \\
    Label projection & 100
\end{tabular}

	\caption{Hyperparameters}
	\label{table:hyperparams}
\end{table*}

\end{document}